# Semantic Visual Localization

Johannes L. Schönberger[1]  Marc Pollefeys[1,3]  Andreas Geiger[1,2]  Torsten Sattler[1]

[1]Department of Computer Science, ETH Zürich  [2]Autonomous Vision Group, MPI Tübingen  [3]Microsoft

{jsch,pomarc,sattlert}@inf.ethz.ch   andreas.geiger@tue.mpg.de

## Abstract

*Robust visual localization under a wide range of viewing conditions is a fundamental problem in computer vision. Handling the difficult cases of this problem is not only very challenging but also of high practical relevance, e.g., in the context of life-long localization for augmented reality or autonomous robots. In this paper, we propose a novel approach based on a joint 3D geometric and semantic understanding of the world, enabling it to succeed under conditions where previous approaches failed. Our method leverages a novel generative model for descriptor learning, trained on semantic scene completion as an auxiliary task. The resulting 3D descriptors are robust to missing observations by encoding high-level 3D geometric and semantic information. Experiments on several challenging large-scale localization datasets demonstrate reliable localization under extreme viewpoint, illumination, and geometry changes.*

## 1. Introduction

Visual localization is the problem of determining the camera pose of one or multiple query images in a database scene. This problem is highly relevant for a wide range of applications, including autonomous robots [58] and augmented reality (AR) [32, 43], loop closure detection [18, 20, 81] and re-localization [38, 51] in SLAM [36, 46], and Structure-from-Motion (SFM) [54, 57] systems.

There are three types of approaches to the localization problem: *Structure-based* methods represent the scene by a 3D model and estimate the pose of a query image by directly matching 2D features to 3D points [11, 37, 51, 66, 80] or by matching 3D features to 3D points, if depth information is available [30, 81]. *Image-based* methods model the scene as a database of images [14, 18, 53, 65, 68]. They use image retrieval techniques to identify the database images most relevant to the query, which are then used to estimate the pose from 2D-3D matches. *Learning-based* methods represent the scene by a learned model, which either predicts matches for pose estimation [7, 8, 59, 69] or directly regresses the pose [29, 70]. This paper follows the structure-based ap-proach and represents the database scene by a semantic 3D map. Given a query image together with its semantic segmentation and depth map, we construct a 3D semantic query map from which we extract local descriptors. Using 3D-3D matches between query and database descriptors, we align the maps to obtain the query pose estimate.

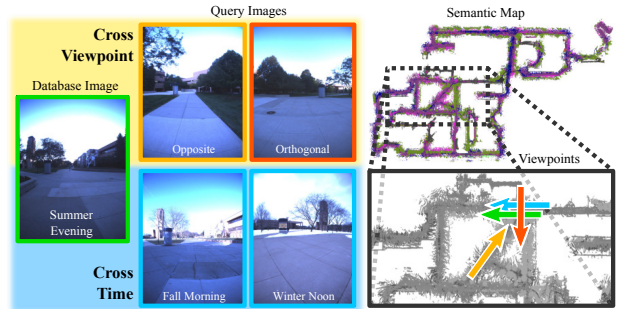

Figure 1: We propose a semantic localization technique which is able to match features over extreme appearance changes across viewpoints and time. In this example, the database contains only images captured in summer and from one particular viewpoint, yet our method correctly localizes images with strong viewpoint, illumination, and seasonal changes.

All of the approaches, including ours, explicitly or implicitly measure the (visual or structural) similarity between a query image and the database scene representation. Thus, they assume that the query and database images depict the scene under sufficiently similar conditions in viewpoint, illumination, and scene geometry. As shown in Fig. 1, these assumptions are easily violated in practice. Different illumination within a single day causes strong variation in appearance while seasonal changes significantly affect the scene geometry. Similarly, strong viewpoint changes lead to severe perspective distortion and often result in little structural overlap between the query and the database. Yet, robustness against such changes is important, e.g., for AR devices or robots to re-localize robustly in a changing environment.

The main challenge in this setting is successful data association between the query and the database. Existing image- and structure-based methods use local features designed to be discriminative, e.g., [30, 60, 61, 81], such that descriptors of the same physical point are close in descriptor space while unrelated points are far apart. However, strong changes in viewing conditions, e.g., in appearance or geometry, demand for an invariant embedding which contradicts the discriminative learning objective of these approaches. In theory, such invariance could be implemented by learning a more complex descriptor comparison function [23,78]. Yet, in practice, such methods do not scale well, as they require an expensive pairwise comparison of descriptors.

To overcome this limitation, we present a novel approach to descriptor learning that is based on a generative rather than a discriminative model. The core idea is to learn an embedding in Euclidean space that retains all information required to recover the scene appearance under different viewing conditions. Our embedding encodes high-level 3D geometric and semantic information and thus allows us to handle strong viewpoint changes as well as moderate changes in scene geometry, e.g., due to seasonal changes. More specifically, we propose to learn a generative descriptor model based on the auxiliary task of 3D semantic scene completion. Given a partially observed scene, the goal of this auxiliary task is to predict the complete scene. A key insight of our paper is that semantics provide strong cues for the scene completion task, resulting in drastically improved descriptors. We show that our descriptors can be learned in a self-supervised manner without explicit human labeling. The learned descriptors generalize to new datasets and different sensor types without re-training.

In summary, this paper makes the following **contributions**: **(i)** We propose a novel approach to visual localization based on 3D geometric and semantic information. **(ii)** We formulate a novel method to the descriptor learning problem based on a generative model for 3D semantic scene completion. The latent space of our variational encoder-decoder model serves as our descriptor and captures high-level geometric and semantic information. **(iii)** We demonstrate the effectiveness of our approach on two challenging problems: Accurate camera pose estimation under *strong viewpoint changes* and *illumination/seasonal changes*. Even without semantics, our approach outperforms state-of-the-art baselines by a significant margin, demonstrating the power of generative descriptor learning in localization. Incorporating semantic information leads to further improvements. To the best of our knowledge, ours is the first approach which reliably estimates accurate camera poses under such challenging conditions. In addition, our method generalizes to new datasets with different types of sensors without re-training.

## 2. Related Work

**Traditional Approaches.** Most existing large-scale localization methods use local features such as SIFT [42] to establish 2D-3D matches between features in a query image and points in a SFM model [37, 38, 51, 66, 80], These correspondences are then used to estimate the camera pose. Descriptor matching is typically accelerated using prioritization [38,51] or efficient matching schemes [39, 43]. Co-visibility information [37, 51], an intermediate image retrieval step [26, 53], and geometric outlier filtering [11, 66, 80] aid in handling ambiguous features arising at large scale. If available, depth information can be used to remove perspective distortion effects before descriptor extraction [74, 79] or to directly extract descriptors in 3D [30, 48, 81]. However, even depth-based approaches fail in the presence of strong viewpoint or appearance changes due to a lack of visual or structural overlap. In contrast, we make our approach more robust to such drastic changes by learning a novel 3D descriptor specifically for these conditions. Recent learning-based methods for visual localization either learn to associate each pixel to a 3D point [7,8,59,69] or learn to directly regress the camera pose from an image [29, 70]. The principal drawback of both approaches is that they need to be retrained for each dataset. In contrast, our learned semantic descriptors generalize across datasets.

**Semantic Localization.** A popular strategy for semantic localization is to focus on features found on informative structures [34, 45] and to re-weight or discard ambiguous features [33]. Similarly, individual features [34] or Bag-of-Words representations [2, 62] can be enhanced by combining local features with semantics as a post-processing step. In contrast, our approach learns to combine semantics and geometry into a single and more powerful descriptor.

An alternative strategy to semantic localization is to use high-level features such as lane markings [58], object detections [3, 4, 49, 67], discriminative buildings structures [72], or the camera trajectory of a car [10]. These approaches need object databases or maps containing the same types of objects, which either requires careful manual annotation [58] or pre-scanning of objects [49]. In addition, the feature extraction and matching process is often a complex and hand-crafted solution tailored to specific objects [3, 4]. In contrast, our model learns a general semantic scene representation in a self-supervised fashion from data, eliminating the need for hand-crafted solutions or manual labeling.

**Descriptor Learning.** The traditional approach to descriptor learning in the general setting is to learn a discriminative embedding in Euclidean space from corresponding 2D patch samples [9, 23, 35, 60, 61]. The embedding function should produce similar descriptors for patches depicting the same physical structure and dissimilar descriptors for unrelated patches. The same approach also applies to 3D voxel volumes [81] and point clouds [30]. Typically, these descriptors are learned for local patches or local volumes in order to handle (partial) occlusions. In contrast, we are interested in learning descriptors with a larger spatial context in order to obtain a more powerful, high-level understand-ing of the scene. Consequently, we learn 3D descriptors for relatively large 3D semantic voxel volumes. The main challenge in our setting is that descriptors in the query and database map only have partial structural overlap due to their large spatial context and due to strong occlusions when matching under extreme viewpoint changes. We thus exploit the auxiliary task of semantic completion [63] to learn an embedding that is invariant to occlusions. In contrast to learning complex matching functions [23, 78] that are expensive to compute, our descriptor is embedded in Euclidean space and can be matched efficiently at large scale.

One application of our approach is localization under illumination and seasonal changes. There exists work on training local [41] or image-level [1, 14, 68] descriptors that are robust under such changes. Due to the challenge of obtaining accurately posed images under different conditions [52], these approaches are trained on data with little viewpoint changes, e.g., from webcams [14]. Thus, such approaches are not very robust under viewpoint variations. In contrast, we explicitly train on data with strong viewpoint changes and demonstrate that our model also generalizes to illumination and seasonal changes.

**Semantic Model Alignment.** The key idea of our method is to use the geometry and semantics to establish correspondences for pose estimation. Thus our approach is also related to methods aligning 3D models through semantic features [15, 16, 67, 77]. Cohen et al. [15, 16] use semantic features to stitch visually disconnected SFM models. Toft et al. [67] use a similar idea for camera pose refinement in localization. Yu et al. [77] use 3D object detections as features in a semantic ICP approach. These approaches use hand-selected semantic features, which are often ambiguous, e.g., there might be multiple cars in the scene. Hence, the association problem is solved either via brute-force search [16] or by assuming an initial alignment [15, 67, 77]. In contrast, our approach learns descriptors with a more general semantic scene understanding that can be matched efficiently at large scale.

**Aerial-Ground Localization.** A related problem to ours is tackled by work on matching ground-level imagery against overhead maps to obtain coarse location estimates under orthogonal viewpoint changes [13, 40, 71, 73]. However, these methods are specific to this problem and cannot be used for accurate ground-level to ground-level localization, which is the focus of this paper.

## 3. Semantic Visual Localization

In this section, we describe our proposed method for semantic visual localization. The input to our system is a set of color images with associated depth maps $\mathcal{I} = \{I_i\}$ and, for database images, their respective camera poses $\mathcal{P} = \{P_i\}$ with $P_i \in SE(3)$. Given the subset of database images $\mathcal{I}_D$ and their camera poses $\mathcal{P}_D$, we create a global 3D semantic map $M_D$ in a pre-processing step. For a query $\mathcal{I}_Q$ of one or multiple images, we compute a local 3D semantic map $M_Q$ and establish 3D-3D matches between $M_Q$ and $M_D$ to determine the unknown query poses $\mathcal{P}_Q$. This localization procedure should be robust to extreme viewpoint and illumination changes between the database and the query images. While lower-level radiometric and geometric information typically varies significantly under different viewpoints and illumination, semantic information is comparatively invariant to these types of transformations through higher-level scene abstraction. This is the main motivation for our proposed semantic visual localization method which comprises the following three steps: In an offline step, we learn robust local descriptors by exploiting semantic scene completion as an auxiliary task. During online operation, we use these local descriptors to establish 3D-3D matches between the query and database map. The matches are then used to estimate an alignment between the two maps, which defines a pose estimate for the query. In the following, we first describe how we construct the 3D semantic maps from which we learn and extract our proposed descriptors. We then explain the proposed descriptor matching and pose estimation stages.

### 3.1. Semantic Segmentation and Fusion

We first compute dense pixelwise semantic segmentations $\mathcal{S} = \{S_i\}$ for all input images, where each pixel of $S_i$ is assigned a semantic class label $l \in \{1, \ldots, L\}$. Next, we fuse the images into semantic 3D voxel maps $M_D$ and $M_Q$ for the database and query images [22, 25]. Each voxel in the semantic 3D maps takes one of $L+2$ labels, i.e., a voxel is either occupied with one of the $L$ semantic classes or it is labeled as free space $L_F$ or unobserved space $L_U$. The task of localization is to find the transformation $P \in SE(3)$ that best aligns a query to the database map.

Given a robust semantic classifier, e.g., trained specifically for different seasons, the semantic maps are inherently invariant to large illumination changes and geometric variations up to the voxel resolution. Note that using semantics, it is easy to determine reliable classes and, e.g., to ignore dynamic objects such as cars. While semantics abstract high-level scene information, large spatial context is needed for an unambiguous, instance-level understanding of the scene. However, a larger spatial context inherently leads to missing observations due to occlusions. For example, in the case of different viewpoints, the volumetric overlap between the query and database maps may be very small. In the extreme case of opposite viewing directions, there might be no structural overlap between the two maps. Hence, one main challenge for our pipeline is to robustly find matches between the query and database maps in the absence of common observations. In the following, we de-scribe how to learn an encoding of the database and query maps $M_D$ and $M_Q$ that is invariant to such missing observations through a semantic understanding of the scene.

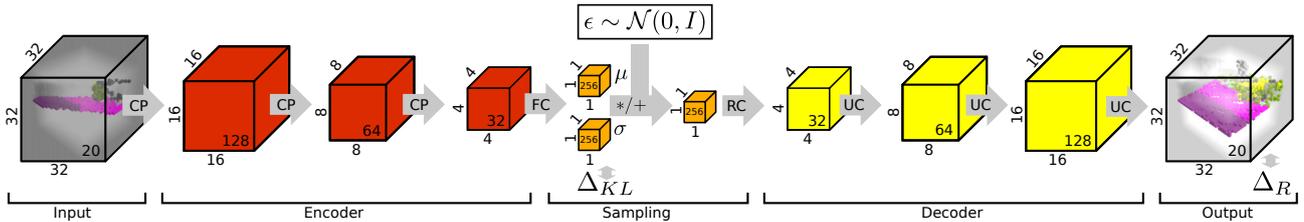

Figure 2: **Variational Encoder-Decoder Architecture.** Legend: CP = Convolution + Pooling, FC = Fully Connected, RC = Reshape + Convolution, UC = Upsampling + Convolution, $\Delta_{KL}$ = KL Divergence wrt. $\mathcal{N}(0, I)$, $\Delta_R$ = Reconstruction Loss. The numbers at the bottom right of each block denote the number of feature channels. The network takes incomplete semantic observations as input (left) and predicts completed semantic subvolumes (right). The latent code $\mu$ forms our descriptor.

### 3.2. Generative Descriptor Learning

The underlying goal of our localization method is to estimate the transformation $P$ from 3D-3D matches between $M_D$ and $M_Q$. Since the query and database maps typically differ in size and coverage, we find correspondences between subvolumes $v_D \in M_D$ and $v_Q \in M_Q$ of size $V^3$. To establish these correspondences, we learn a function that recognizes similar subvolumes. This function should be invariant to missing observations due to the relatively large size of the subvolumes, different viewpoints and moderate geometric deformations between the query and database map, dynamic objects in the scene, sensor noise, etc. In particular, the function should identify the same object even when seen from different viewpoints and under different illumination. We will show that semantic scene understanding is key to learning such an invariant function.

The two traditional approaches to solving this problem are to either learn a matching function $f(v_D, v_Q)$ [23, 78] or an embedding $f(v)$ [30, 81]. The latter approach aims to find an encoding function that maps the same subvolumes to similar points in (Euclidean) space. While a learned matching function in theory has more discriminative power, it also imposes high computational cost as it requires exhaustive pairwise comparisons, which is intractable at large scale. Thus, we learn an embedding that is evaluated only once per subvolume rather than per pair of subvolumes.

More concretely, we learn an encoding function $f(v) \in \mathbb{R}^N$ that maps a subvolume to a lower-dimensional descriptor which jointly encodes the scene semantics and geometry. To recognize the same object from different or even opposing viewpoints, this encoding must contain enough information to hallucinate the unobserved parts of the subvolume. Towards learning such a robust encoding, we define the auxiliary task of semantic scene completion. This auxiliary task is described by the function $h(v)$ that hallucinates the geometry and the semantics of the unobserved parts of its input. We use a 3D variational encoder-decoder $h(v) = g(f(v))$, where $f$ is a neural network which encodes the incomplete subvolume and $g$ is a neural network which hallucinates the complete subvolume. To learn the distribution of the space of subvolumes and to ensure that the same physical subvolumes map to nearby encodings in Euclidean space, we enforce a Gaussian prior on $(\mu, \sigma) = f(v)$ using variational sampling. Our formulation is similar to the original variational auto-encoder [31] with the difference that we encode an incomplete sample and decode the complete sample.

For learning $h$, we generate training data using volumetric fusion. We first fuse all training images $\mathcal{I}_T$ into a volumetric representation. This yields a nearly *complete* representation $M_T$ of the scene. In addition, we create *incomplete* volumetric representations $M_{T_i}$ for each image $I_{T_i} \in \mathcal{I}_T$ individually. During stochastic gradient descent, we randomly sample incomplete subvolumes $\bar{v}$ in $\{M_{T_i}\}$ and find its corresponding complete subvolume $\hat{v}$ in $M_T$. The task of $h$ is to denoise the observed parts and to hallucinate the unobserved parts of the incomplete subvolume. The learning objective is the semantic reconstruction loss $\Delta_R = \mathrm{E}(h(\bar{v}), \hat{v})$ using the categorical cross entropy measure $E(\cdot, \cdot)$. Together with the Gaussian prior, the overall training objective is defined as $\Delta = \Delta_R + \Delta_{KL}$ where $\Delta_{KL}$ measures the Kullback-Leibler divergence between the latent code $f$ and $\mathcal{N}(0, I)$. The architecture of $h$ is illustrated in Fig. 2 and examples are shown in Fig. 3.

We learn the model for a fixed voxel size using the same orientation for the incomplete and complete subvolumes $\bar{v}$ and $\hat{v}$. For additional data augmentation and robustness to noise, we jointly rotate the subvolumes using random orientations and perturb the occupancy of the incomplete subvolumes using dropout [64]. Note that no human labeling is required since we employ pre-trained semantic classifiers for this task. As described next, our semantic localization pipeline only uses the encoder part $f$ of the full model $h$.

### 3.3. Bag of Semantic Words

The previous section described how to learn a discriminative function $f$ that maps semantic subvolumes $v$ to low-dimensional latent codes $\mu$. We use this function to create a *Bag of Semantic Words* that encodes the semantic 3D scene layout of a map. The bag of semantic words $\mathcal{F}(M) = \{f(v_j)\}$ with $j = 1\ldots|M|$ is defined as the set of descriptors $f(v_j)$ computed for all occupied subvolumes $v_j$ in $M$. We consider a subvolume to be occupied if at least one voxel within the subvolume (not necessarily the center voxel) is occupied. Note that given an *incomplete* map, the bag of semantic words is a description of its *complete* semantic scene layout. Our localization pipeline uses this representation to robustly match the query map to the database map, as detailed in the following.

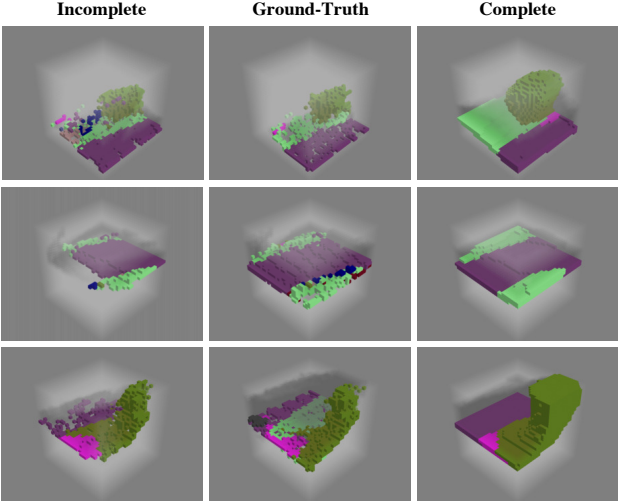

Figure 3: Example input $\bar{v}$ and output $h(\bar{v})$ from the KITTI dataset for our semantic completion auxiliary task. The incomplete input is completed using our encoder-decoder network $h$, while the multi-view fusion $\hat{v}$ is the ground-truth.

### 3.4. Semantic Vocabulary for Indexing and Search

We establish correspondence between subvolumes in the query and database using nearest neighbor search in the descriptor space $f(v) \in \mathbb{R}^N$ using the Euclidean metric. For efficient semantic word matching, we build a semantic vocabulary [47] in an offline procedure using the bag of semantic words of the training dataset. We quantize the space of descriptors $\mathcal{F}(M_T)$ using hierarchical k-means and a $N_B$-dimensional Hamming embedding [27]. We index all semantic words of the database map $\mathcal{F}(M_D)$ into the resulting vocabulary tree, which serves as an efficient data structure for matching. To find matches between a given query image and the database, we find the top $K = 5$ nearest database words $\mathcal{D}$ for each query word $f(v_j) \in \mathcal{F}(M_Q)$ by traversing the vocabulary tree and finding nearest neighbors in Hamming space. Since our descriptors are *rotation variant*, as they are trained on aligned subvolumes (see Section 3.2), and, generally, we have no *a priori* knowledge about the orientation of the query, we perform the same query for a fixed set of orientation hypotheses $\theta \in SO(3)$ while the database remains fixed. The set of putative matches $\mathcal{D}(\theta)$ for the different orientations provide evidence for the location of the query. The next section details how to accurately localize the query based on this evidence using a joint semantic map alignment and verification.

### 3.5. Semantic Alignment and Verification

Given the putative matches $\mathcal{D}(\theta)$ from Section 3.4, we seek to find the transformation $P \in SE(3)$ that best aligns the query to the database map. Specifically, a good alignment is established if both the geometry (i.e., occupancy) as well as the semantics agree. Due to the rotation variance of our descriptors, a single 3D-3D match between the query and the database defines a transformation hypothesis $P$, which is composed of the rotation defined by $\theta$ and the translation $t \in \mathbb{R}^3$ defined by the spatial offset of the corresponding subvolumes. We exhaustively enumerate all transformation hypotheses defined by the matches. To verify a single transformation hypothesis, we then align the query to the database map using $P$ and count the number of correctly aligned voxels of the query map. A correctly aligned voxel matches both in terms of geometry and semantics, i.e., an occupied voxel in the aligned query map must also be occupied in the spatially closest voxel in the database map. In addition, the spatial distance of their voxel centers must be smaller than $\kappa$ and their semantic class labels must match exactly. We ignore unobserved voxels in both the query and the database map. To further refine the alignment, we use the iterative closest point algorithm [6], where closest points are defined as the set of correctly aligned voxels in the previous iteration. Finally, we rank the transformation hypotheses by the ratio $\tau$ of correctly aligned over occupied voxels in the query map. The top-ranked hypotheses define the query pose estimates as the output of our system.

## 4. Experiments

In this section, we compare our method to the state-of-the-art techniques on several large-scale localization benchmark datasets. The following sections explain the setup and results of our experiments in detail.

### 4.1. Datasets

**KITTI.** We evaluate the localization performance in the setting of extreme viewpoint changes on the KITTI odometry dataset [21] comprising 11 sequences with ground-truth poses. 6 of these sequences contain loops with extreme viewpoint changes. First, we evaluate on the traditional loop closure scenario including all images independent of viewpoint change. For this, we construct the database map from all images, while the query map is constructed from a single image. In addition, we also perform experiments when the database only contains images from significantly different viewpoints (90° or 180°) and under different ap-pearance (Fig. 4). In this case, all images with similar viewpoint are excluded from the database.

**NCLT.** In addition, we use the NCLT dataset [12] to evaluate the localization performance under extreme appearance changes caused by short-term illumination changes over several hours and long-term seasonal changes over several months. The dataset was acquired biweekly during 1.5 hour sessions over the course of 15 months and we selected 4 sequences that span the different modalities of daytime (morning, midday, afternoon, evening), weather (sunny, partially cloudy, cloudy), (no) foliage, and (no) snow.

### 4.2. Setup

For all evaluations, we compute gravity-aligned semantic maps by using either the integrated inertial sensors (NCLT) or through vanishing point detection in the images (KITTI). This reduces the space of orientation hypotheses $\theta$ to the rotation around the gravity axis. We use the raw output of an off-the-shelf semantic classifier [76] trained on the Cityscapes [17] dataset. This classifier segments the scene into $L = 19$ semantic classes and we only consider the maximum activation per pixel and discard any pixels with *sky* labels. We adapted the volumetric fusion approach by Hornung et al. [25] using (multi-view) stereo depth maps [24] (KITTI) and sparse LIDAR measurements (NCLT) for efficient large-scale semantic fusion. We extract subvolumes $v$ of size $32^3$ at a fixed voxel resolution of 30cm, resulting in a $10m^3$ spatial context. At this resolution, the bag of semantic words for a single image query map contains several thousand descriptors for a fixed set of 18 uniformly spaced orientation hypotheses $\theta$. A single forward pass of one volume takes around 1ms on a NVIDIA Titan X GPU while the geometric verification has negligible performance impact in comparison, leading to an average throughput of around one query image per second, which is on par with the fastest baselines we compare to. For pose estimation, we empirically choose a maximum error of $\kappa = 3$ meters and a minimum overlap ratio of $\tau = 0.3$.

### 4.3. Training

Throughout all experiments, we employ the same semantic descriptor and vocabulary trained offline on two sequences of around 10k and 20k frames [75] with accurate ground-truth poses acquired over several kilometers. The dataset contains a large number of loops with extreme viewpoint changes and is independent of the evaluation datasets. Note that this dataset is much more similar to KITTI as compared to NCLT, since it is an autonomous driving dataset with a stereo camera pair that we use for volumetric fusion. Nevertheless, we show that our descriptor generalizes well to NCLT. We train our encoder-decoder architecture on 16M random subvolumes using SGD and choose a latent code size of $N = 256$ as a trade-off be-tween speed and accuracy. In our experiments, $N < 64$ led to significantly reduced reconstruction and localization performance, while $N > 512$ did not significantly improve the results. Our semantic vocabulary is represented by $2^{16}$ semantic words embedded in a $N_B = 64$ dimensional Hamming space, using a hierarchical branching factor of 256.

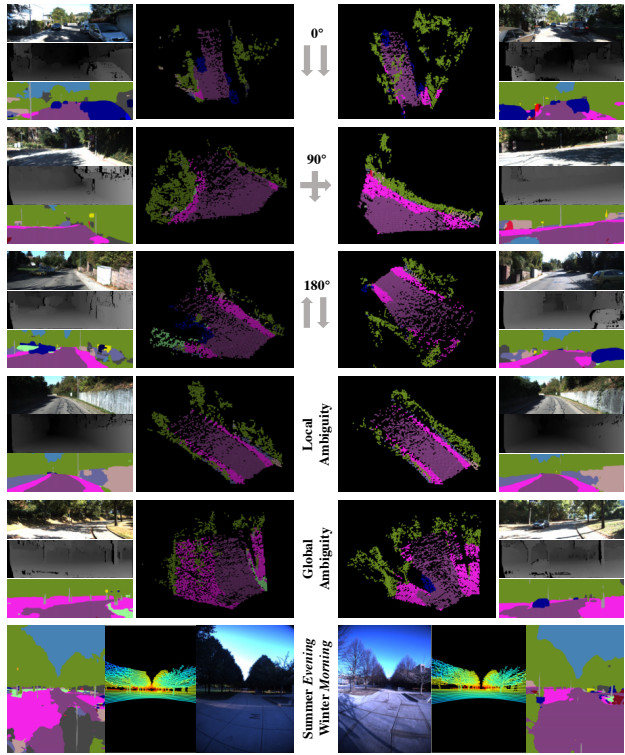

Figure 4: Example scenes in the KITTI and NCLT datasets for the different loop closure scenarios, including two failure cases caused by local and global ambiguities.

### 4.4. Baselines

In the following, we briefly present the chosen state-of-the-art baselines for our evaluation. We evaluate the localization performance on the ratio of correctly localized query images within a given error threshold. In addition, we show the rank-recall curves for an error threshold of 1m and 5m. For additional implementation details of the baselines, we refer the reader to the supplementary material.

**SIFT and DSP-SIFT.** We employ a state-of-the-art visual localization pipeline [50, 56] using SIFT [42]. This pipeline is based on a visual vocabulary tree embedded in a Hamming space [27] with visual burstiness weighting [28] and uses 2D-2D matching on the top-ranked retrievals to obtain 2D-3D correspondences for absolute pose estimation. Instead of SIFT, we also evaluate using DSP-SIFT [19], which has been shown to perform significantly better [5, 55].

**MSER and VIP.** Furthermore, we replaced the standard

SIFT keypoint detector with MSER [44] features, which are designed for wide baseline matching. Then, we extracted DSP-SIFT features and kept the rest of the SIFT pipeline as described above. In addition, we experimented with Viewpoint Invariant Patches [74], where we rectified the ground plane to cancel the effect of perspective distortion before extracting SIFT features. However, we found that we could not match features for $90°$ and $180°$ viewpoint changes on the KITTI dataset. Our main insight from this experiment was that the geometric and radiometric distortions are too severe for low-level appearance matching.

**DenseVLAD.** To study the impact of image retrieval on the localization performance, we replaced the vocabulary tree based ranking with DenseVLAD [68] as a global image descriptor and otherwise use the SIFT pipeline as described. DenseVLAD produces state-of-the-art retrieval results under large viewpoint [53] and illumination changes [68].

**FPFH, CGF, and 3DMatch.** FPFH [48], CGF [30], and 3DMatch [81] are state-of-the-art hand-crafted and learned geometric shape descriptors used for point cloud matching. We use them as a replacement for our descriptor while keeping the rest of the pipeline as proposed. Equivalent to our setup, we densely extract these shape descriptors using the same spatial descriptor radius and train a custom shape vocabulary on our training dataset. In addition, we fine-tune 3DMatch on our data using corresponding incomplete and complete volumes, resulting in a small performance boost.

**PoseNet and DSAC** We train separate PoseNet [29] models for each sequence as a state-of-the-art representative of end-to-end pose regression methods. We also experimented with DSAC [8] but failed to obtain meaningful results even for the smallest KITTI sequence.

**Ours.** We compare the above state of the art against our semantic localization pipeline, denoted as *Ours (semantic)*. To quantify the performance impact of semantic information on the localization task, we additionally train a version of our descriptor and vocabulary using uniform semantics with only a single label. As a result, the fused 3D maps contain only occupied, free, and unobserved space labels. Opposed to the FPFH, CGF, and 3DMatch, this can be seen as an occlusion-aware geometric shape descriptor, denoted as *Ours (geometric)*. In addition, we show results for fusing multiple frames into the query map, denoted as *Ours (acc.)*. Opposed to fusing a single frame, we fuse five consecutive frames into a query map and therefore demonstrate the benefit of accumulating more evidence over time to obtain less noisy and more complete query maps.

### 4.5. Results

**Scene Completion.** Fig. 3 shows results of the semantic completion task. We attain an average semantic reconstruction accuracy of 87% on the test data. In most cases, the completed volume is a spatially smoothed approximation of the ground-truth. We conclude that the network learns powerful semantic representations of the world and meaningfully hallucinates the missing parts. Using this model for all further evaluations, we next demonstrate the performance of our approach on the task of localization.

**$0°$ Localization Scenario.** First, we evaluated our method on the traditional loop closure scenario when images from similar viewpoints as the query image are indexed in the database. Fig. 5 shows the results for KITTI in the $0°$ column and on the main diagonal for NCLT. Our method achieves state-of-the-art errors for the top-ranked localization proposals and clearly outperforms the other geometric shape descriptors, whereas our semantic and geometric descriptors perform roughly on par. While (DSP-)SIFT and DenseVLAD achieve lower performance for the top-ranked proposals, they clearly outperform all other methods when retrieving many images. This is not surprising as this task is rather easy. Both the query and the database images depict the scene from similar viewpoints and are taken close to each other in time. Thus, low-level image statistics encoded by, e.g., SIFT or DenseVLAD are very discriminative.

**$90°$ Localization Scenario.** To evaluate the methods in a more challenging setting, we consider $90°$ trajectory intersections, which is a common scenario in most robotic applications, e.g., when a car passes a street crossing twice but in orthogonal directions. For this experiment, we excluded all images from the database that are not within a viewpoint change of $90° \pm 20°$ wrt. the query images. The query consists of a short sequence of 20 images and the results are averaged over all 57 intersection cases in the dataset. Fig. 5 shows that this is indeed a very difficult scenario.

As expected, SIFT is not able to localize any query images due to the extreme viewpoint change, while DSP-SIFT and DenseVLAD perform slightly better. In contrast, FPFH and our method are able to localize a significant number of queries due to their invariance to appearance changes. Surprisingly, CGF and 3DMatch are not as robust as FPFH. Moreover, due to the different viewpoints, the geometry between the query and database map is not the same. Attributed to this fact, our approach achieves much better performance than existing geometric descriptors, because ours is specifically learned for invariance against missing observations. Moreover, our semantic descriptor outperforms our geometric descriptor by a large margin, highlighting the importance of semantic information for recognizing scenes from different viewpoints. We obtain another significant boost in performance by accumulating evidence over a short window of five frames.

**$180°$ Localization Scenario.** Equivalent to the previous experiment, we also detected $180°$ trajectory intersections, which, for example, occur when a car passes the same street but in opposite directions. Again, we excluded all images from the database that are not within a viewpoint change of 180° ± 20° wrt. the query images. The results in Fig. 5 are averaged over all such cases. In this scenario, there is almost no visual overlap between the query and the database apart from, e.g., the street or walls. Consequently, SIFT is not able to localize any of the query images, while the geometric approaches succeed in this task. Notably, this task seems to be easier than the 90° scenario. We attribute this to the fact that in KITTI the 180° intersections typically have larger structural overlap between the query and database as compared to the 90° intersections.

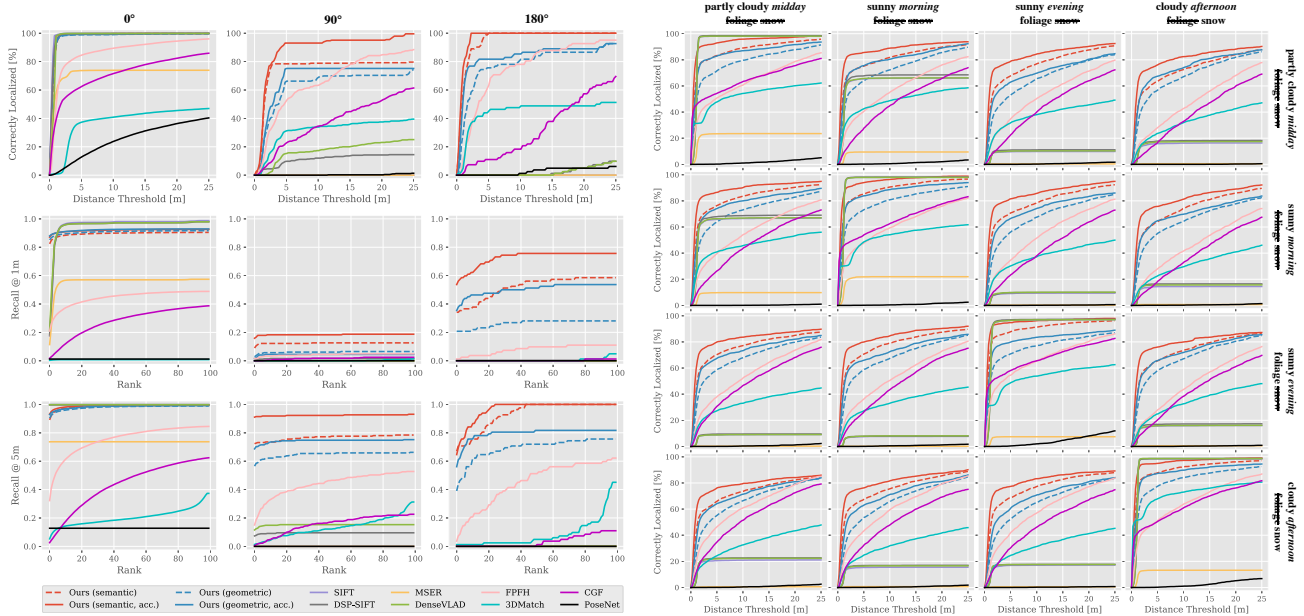

Figure 5: Localization results for the cross-viewpoint (0°, 90°, 180°) and cross-time localization scenarios in the KITTI odometry (left) and NCLT (right) datasets. Striked-through text denotes the absence of a property. Higher is better. See text for details.

**Cross-Time Localization.** The experiments on NCLT show that all but our method fail to robustly localize under extreme appearance changes caused by different geometry (foliage, snow) and illumination (time of day, weather). We observe that, in the cases where (DSP-)SIFT and DenseVLAD succeed, they mostly use stable features on buildings rather than vegetation or ground. In contrast, FPFH and 3DMatch are more robust to illumination changes. However, our method consistently outperforms all other methods across all scenarios. Note that our descriptor was trained on a KITTI-like dataset using stereo for map fusion instead of LIDAR used in NCLT. Furthermore, the same semantic classifier is employed for NCLT and KITTI. This demonstrates that our approach is robust to different types of input data and can be deployed in a wide range of settings without re-training. Looking at Fig. 3, it is not surprising that our approach generalizes to this new task as the network mainly focuses on the overall scene geometry and semantics.

**Failure Cases.** While our method robustly localizes queries in all scenarios, we observed a few common failure cases. The two most systematic errors are joint semantic and geometric ambiguities at a local or global scale (Fig. 4), also commonly occurring in traditional approaches [37, 50]. Local ambiguities are caused by repetitive structures, causing wrong localization in the order of tens of meters. Global ambiguities are more rare and typically result in a localization error of hundreds of meters. Accumulating evidence over multiple frames significantly reduces their impact. Furthermore, our classifier is trained on Cityscapes which only depicts daytime images during spring/summer/fall. Training classifiers tailored to different modalities should further improve the performance of our approach.

## 5. Conclusion

In this paper, we proposed a novel method for localization using a joint semantic and geometric understanding of the 3D world. At its core lies a novel approach to learning robust 3D semantic descriptors. Being the first to demonstrate reliable loop closure and localization even under extreme viewpoint and appearance changes, we believe that our method is an important step towards robust, life-long localization in applications such as autonomous robots or AR. While this paper focused on static scenes, an avenue for future research is the exploration of robustness against strong geometric changes caused by scene dynamics.

**Acknowledgements** This project received funding from the European Union's Horizon 2020 research and innovation program under grant No. 688007 (TrimBot2020).

# Supplementary Material for Semantic Visual Localization

Johannes L. Schönberger[1]  Marc Pollefeys[1,3]  Andreas Geiger[1,2]  Torsten Sattler[1]

[1]Department of Computer Science, ETH Zürich   [2]Autonomous Vision Group, MPI Tübingen   [3]Microsoft

{jsch,pomarc,sattlert}@inf.ethz.ch   andreas.geiger@tue.mpg.de

In this supplementary document, we first give additional implementation details of the proposed semantic visual localization pipeline. Next, we show several examples of semantic query and database maps alongside the obtained localizations and correspondences. Moreover, we provide additional examples of loop closure success and failure cases.

## 1. Implementation Details

**Ours.** We use a batch size of 32 to train our encoder-decoder network for 2,000 epochs using ADADELTA [4] as an adaptive learning rate method for stochastic gradient descent. We set the initial learning rate to $\eta = 1$ without decay and set the hyperparameters to $\rho = 0.95$ and $\epsilon = 10^{-8}$. All convolutional layers use a filter size of $3 \times 3 \times 3$ using zero-padding and *ReLU* activation followed by a $2 \times 2 \times 2$ max-pooling layer. The fully-connected layers are followed by a *tanh* activation function. Upsampling is implemented by repeating the data in the spatial domain by a factor of $2 \times 2 \times 2$. The final convolutional layer of the decoder is followed by a softmax activation. There is a total of around one million learned parameters in our network. For data augmentation, we apply a dropout of 10% on the voxels of the incomplete volume. The reconstruction loss $\Delta_R \in \mathbb{R}$ is emphasized by a factor of 10 relative to the Gaussian prior loss $\Delta_{KL} \in \mathbb{R}$. In addition, for faster convergence during training, the reconstruction loss on occupied voxels is emphasized by a factor of 10. In the experiments, we use 18 uniformly spaced orientation hypotheses $\theta = \{0°, 20°, \ldots, 340°\}$ around the gravity axis.

**SIFT.** For SIFT feature detection, we use 4 octaves starting with a two times up-sampled version of the original image, 3 scales per octave, a peak threshold of $\frac{0.02}{3}$, an edge threshold of 10, and, due to the gravity-aligned input, an upright orientation assumption. The visual vocabulary is represented by $2^{16}$ visual words embedded in a $N_B = 64$ dimensional Hamming space and using a hierarchical branching factor of 256. Using these settings, we obtain several thousand descriptors per image. Localization is performed using a traditional image retrieval setup with two-view geometric verification on the top-ranked retrievals and 2D-3D camera pose estimation inside RANSAC followed by a non-linear refinement. A camera pose is considered as verified, if it has at least 15 2D-3D inlier correspondences. The 3D map is obtained through fusion of all database depth maps. Similar setups achieve state-of-the-art results [2, 3].

**DSP-SIFT.** We use the same feature detector as for standard SIFT and a total of 10 pooling scales uniformly spaced between $\frac{1}{6}$ and 3. We train a new visual vocabulary for nearest neighbor search. Otherwise, we use the same setup as for standard SIFT.

**MSER.** Using the same setup as for DSP-SIFT, we replaced the SIFT keypoint detector with MSER using a step size between 2 intensity threshold levels, region sizes between 30 and 14000 pixels varying by a maximum of 25%. We extract DSP-SIFT descriptors as described previously for the detected regions and train a new visual vocabulary for nearest neighbor search.

**VIP.** For our VIP experiments, we manually selected road regions in multiple images from different viewpoints, fitted a plane through the 3D road points, normalized image to a fronto-parallel viewpoint, densely extracted SIFT descriptors, and matched them exhaustively between the images. For all but very similar viewpoints, we failed to establish correct correspondences between the images. Our main insight from this experiment was that the geometric and radiometric distortions are too severe for low-level appearance matching. We therefore excluded VIP from the further evaluation.

**DenseVLAD.** Using the same setup as for DSP-SIFT, we extract a 4096 dimensional global image descriptor using DenseVLAD, which replaces the visual vocabulary based image retrieval pipeline. To find nearest neighbor images for a given query images, we exhaustively compare the global image descriptors from the query to the database and then perform two-view geometric verification on the top-ranked retrievals, equivalent to the DSP-SIFT experiment.

**FPFH.** Using the same keypoint locations and geometric verification approach as for our method, we extract 33 dimensional FPFH descriptors and train a new vocabulary for nearest neighbor search.

**CGF.** Equivalent to FPFH, we replaced our learned descriptors with 32 dimensional CGF descriptors and train a new vocabulary for nearest neighbor search. We consistently oriented the point cloud normals between the query and database maps towards the cameras.

**3DMatch.** For this experiment, we tried both the pre-trained 3DMatch models and also fine-tuned the descriptor using corresponding complete and incomplete subvolumes that we also used to train our descriptor. The fine-tuned model performs slightly better and we use it for our experiments. Equivalent to FPFH, we then replaced our learned descriptors with 512 dimensional 3DMatch descriptors and train a new vocabulary for nearest neighbor search.

**PoseNet.** For each database, we train a separate PoseNet model from scratch until convergence, which required more than 2 days of training for the largest models. We then regress the pose for each query image, which serves as the single, top-ranked pose hypothesis for the evaluation.

**DSAC.** For this experiment, we trained DSAC from scratch using the suggested initialization protocol, which took around 2 days for the smallest KITTI odometry sequence 04. However, we could not produce meaningful pose estimates and our main insight from investigating the issue was that the current DSAC approach has problems with repetitive structures and larger scale outdoor scenes. We therefore excluded DSAC from the further evaluation.

## 2. Localization Results

**KITTI** Figs. 1, 2, and 3 visualize localization results for three KITTI odometry sequences. The semantic maps have been built by fusing all images, depth maps, and semantic segmentations of the left camera in a sequence. The depth maps were computed by two-view stereo between the left and right camera using semi-global matching [1]. The images, depth maps, and semantic segmentations are jointly fused into semantic 3D maps, which are stored in an efficient Octree data structure at a maximum leaf node resolution of 0.3m. Visualized are all leaf nodes and their corresponding most likely semantic class labels. In addition, we show one example of the different loop closure scenarios and two hard cases caused by ambiguities. For the $90°$ scenario, we excluded all images from the database that are not within a viewpoint change of $90° \pm 20°$ w.r.t. the query image. Equivalently, for the $180°$ scenario, we excluded all images from the database that are not within a viewpoint change of $180° \pm 20°$ w.r.t. the query image. Local ambiguities arise when there are multiple repeating structures in close vicinity. Global ambiguities are rarer and are caused when different parts of the map looks similar both in terms of geometry and semantics. Note that despite ambiguous correspondences between query and database map, our proposed alignment and verification procedure is almost always able to determine the correct location of the query. Fig. 4 shows several alignments between the query and database map, which were obtained using our localization pipeline. Note that the query is aligned accurately even in the case of missing observations and noise.

**NCLT** Figs. 6 and 7 show localization results for the NCLT dataset. Opposed to KITTI we use the LIDAR point cloud and camera 5 of the Ladybug rig for 3D semantic fusion. We use the same descriptor trained on a KITTI-like dataset and, otherwise, use the same setup as for KITTI, demonstrating that our method generalizes across different scene types and different sensors without re-training. Note that NCLT contains both extreme seasonal/illumination changes as well as extreme viewpoint changes between the different datasets. Our method is able to localize robustly in this challenging scenario.

## 3. Local Feature Correspondences

Fig. 5 visualizes corresponding volumes between the query and database maps. The correspondences were established through nearest neighbor search using our proposed semantic descriptor and vocabulary. Note that our descriptor is robust to missing observations and significant noise arising from inaccuracies in depth estimation, semantic segmentation, and 3D fusion, which form the input to our method.

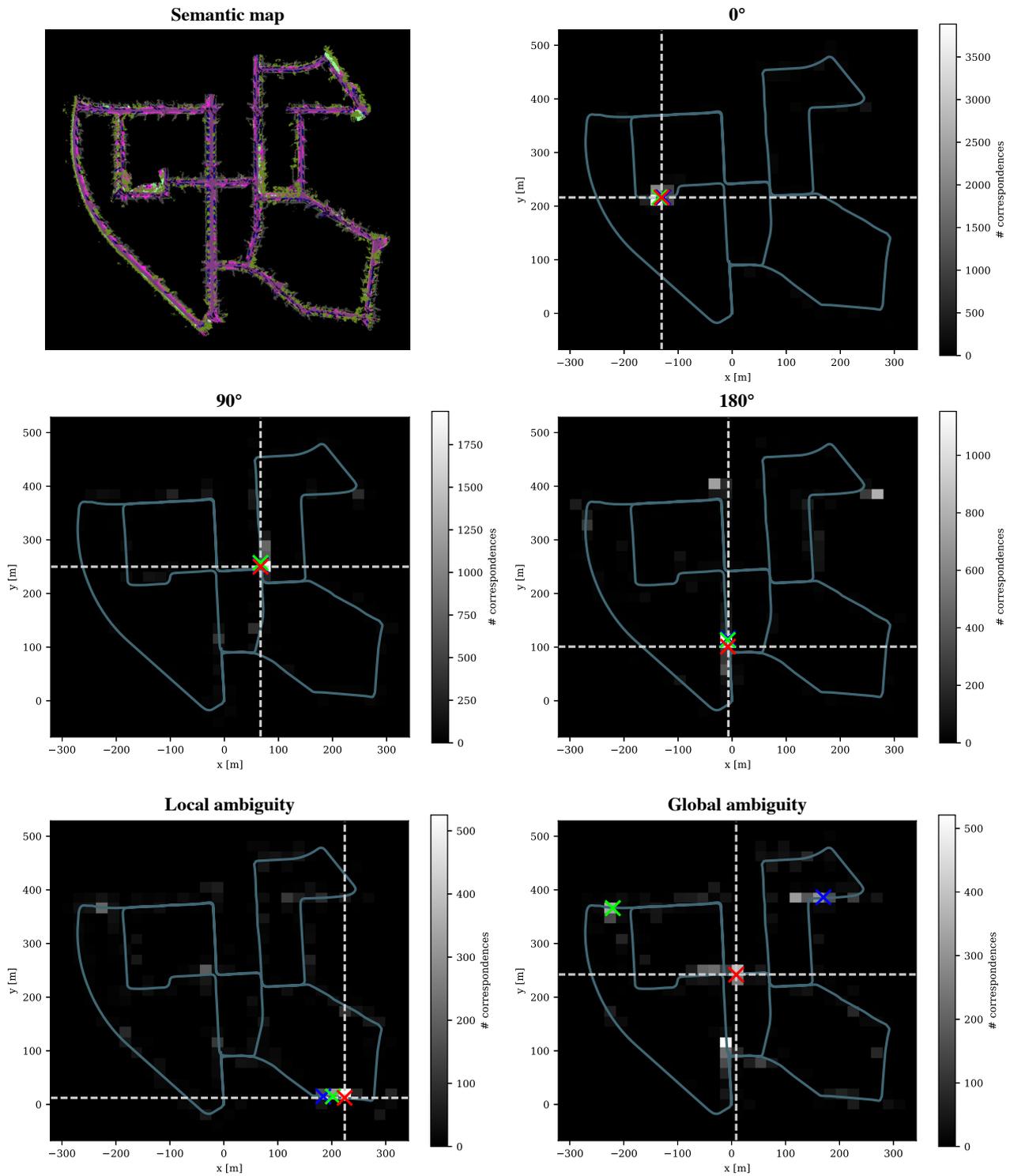

Figure 1: Localization results for sequence `00` of the KITTI odometry dataset. Top-ranked localization results visualized with crosses (red: 1st, green: 2nd, blue: 3rd). Ground-truth location visualized with dotted line. The background histogram visualizes the distribution of corresponding volumes in the database map. Note that, even in the presence of ambiguities, our spatial verification is able to localize correctly.

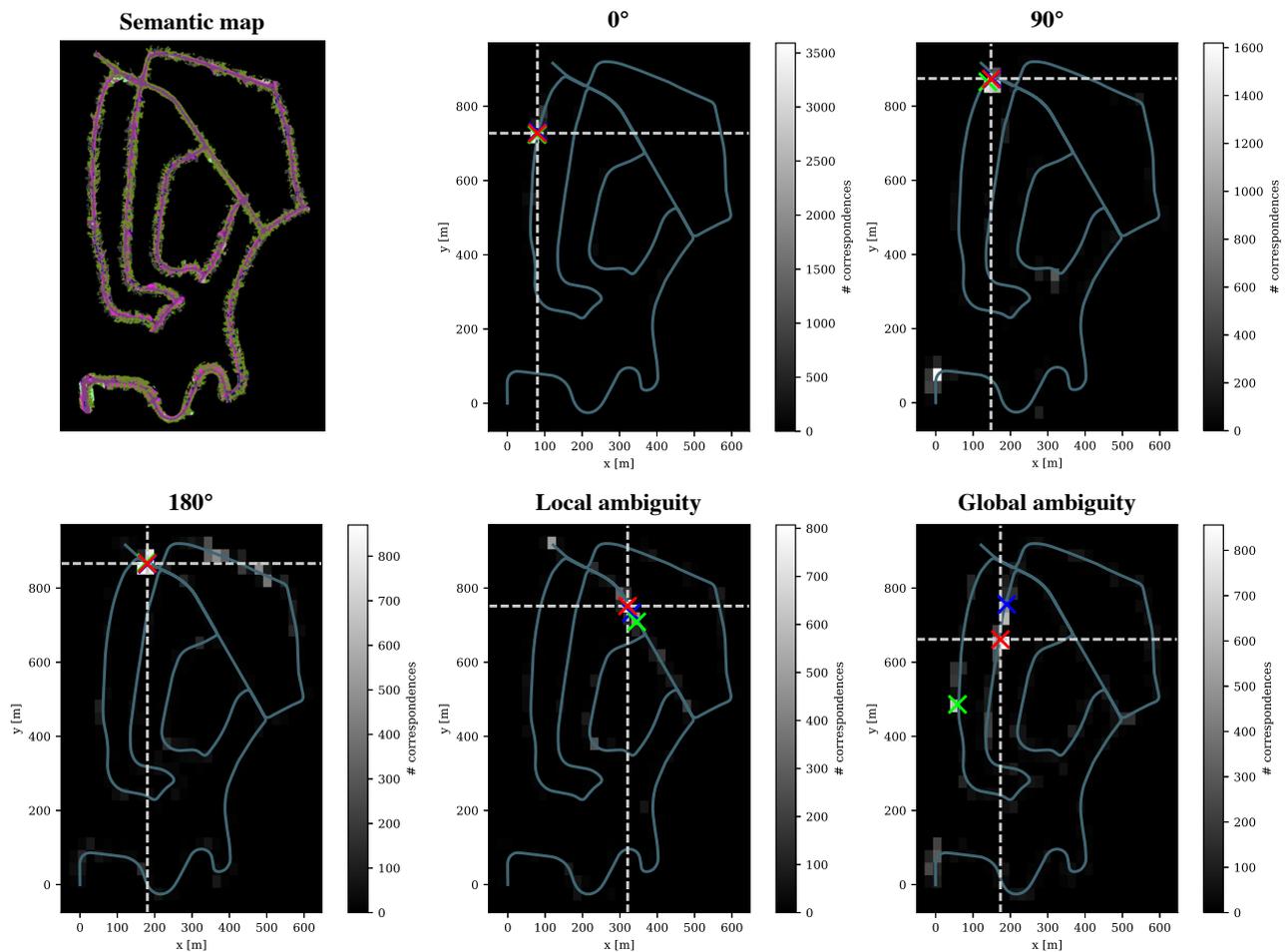

Figure 2: Localization results for sequence `02` of the KITTI odometry dataset. Top-ranked localization results visualized with crosses (red: 1st, green: 2nd, blue: 3rd). Ground-truth location visualized with dotted line. The background histogram visualizes the distribution of corresponding volumes in the database map. Note that, even in the presence of ambiguities, our spatial verification is able to localize correctly.

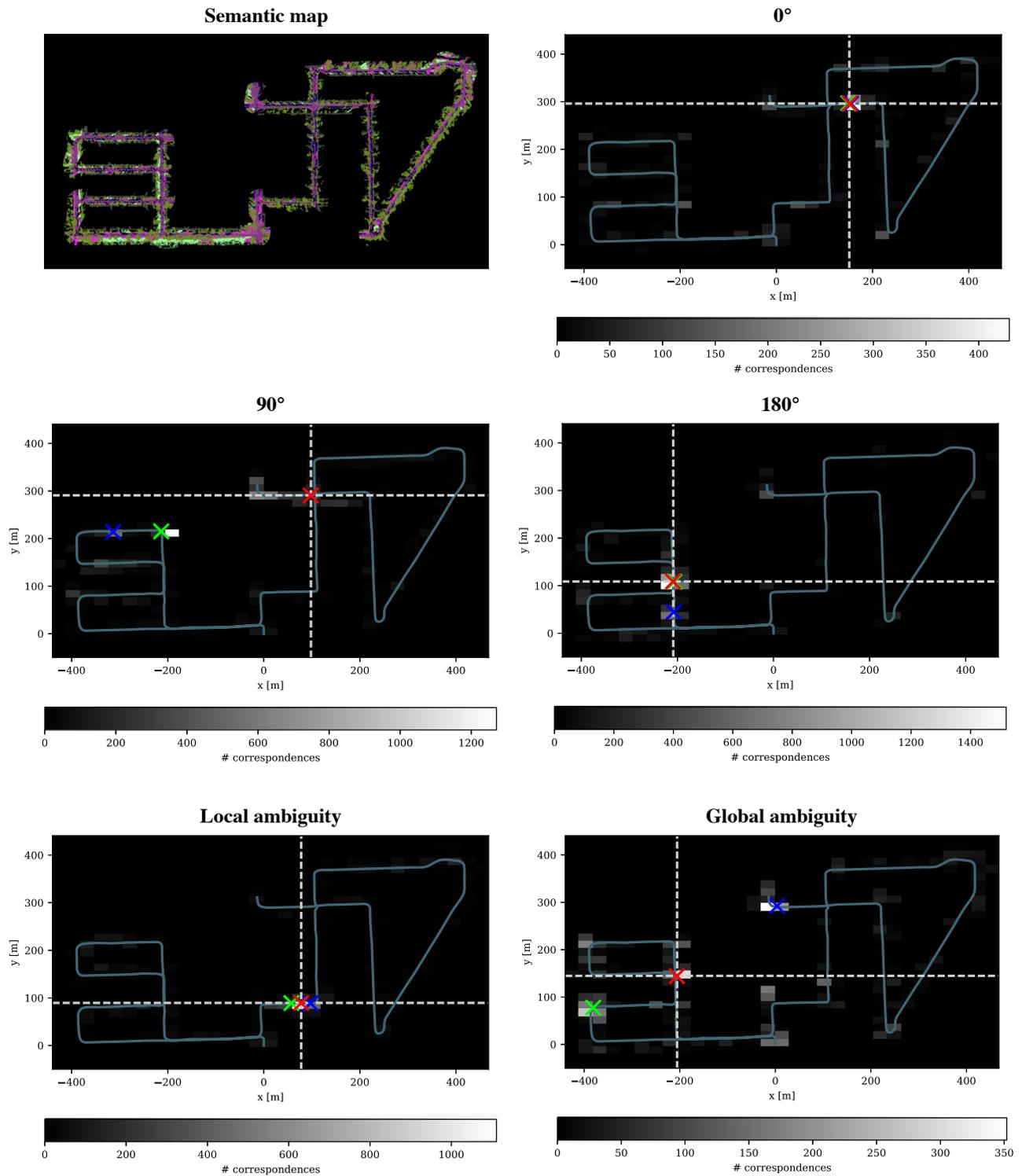

Figure 3: Localization results for sequence `08` of the KITTI odometry dataset. Top-ranked localization results visualized with crosses (red: 1st, green: 2nd, blue: 3rd). Ground-truth location visualized with dotted line. The background histogram visualizes the distribution of corresponding volumes in the database map. Note that, even in the presence of ambiguities, our spatial verification is able to localize correctly.

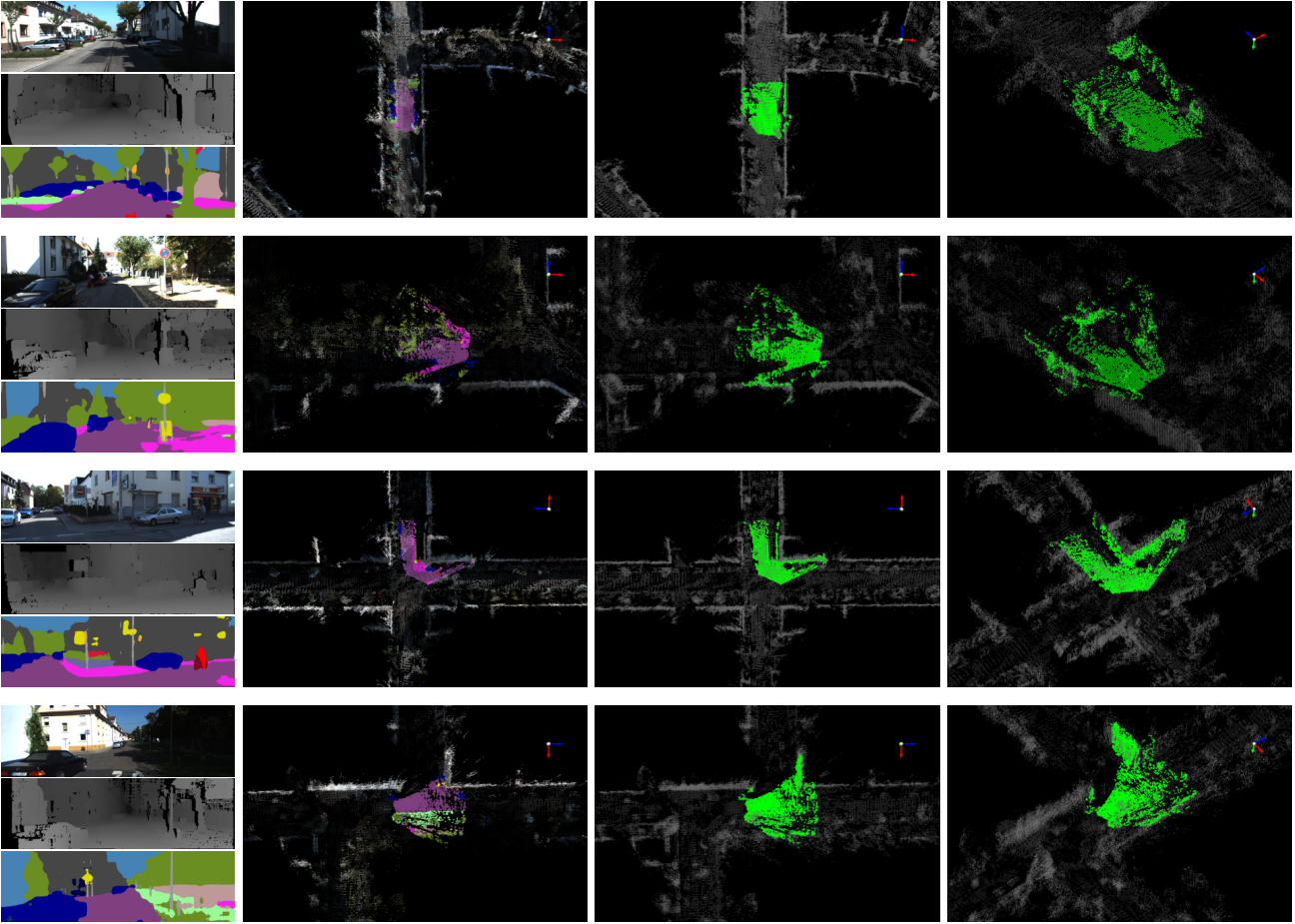

Figure 4: The left column shows the input image and its corresponding depth map and semantic segmentation. The second column shows a top-down view of the aligned query and database map using semantic coloring for the query and RGB coloring for the database. The third column shows the same view but using green color for the query and gray for the database map, while the last column shows the same but from a different viewpoint. Note that the query and database maps are aligned accurately even in the presence of missing observations and significant noise.

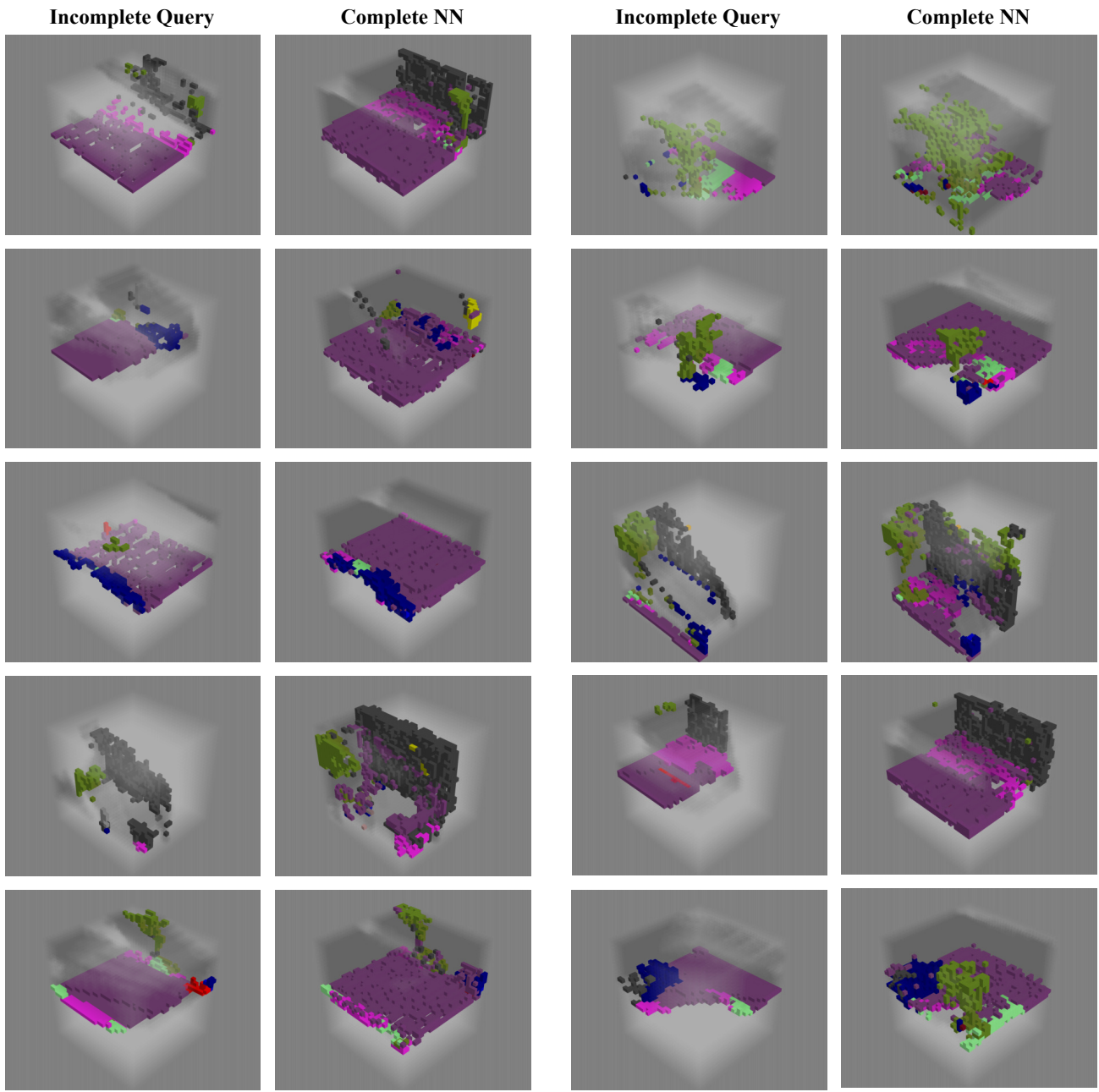

Figure 5: The corresponding volumes in the query map (incomplete query) and retrieved nearest neighbor volumes (complete NN) in the database map. Nearest neighbor search was performed with our proposed semantic vocabulary tree.

| Database | Successful Queries |
|---|---|
| 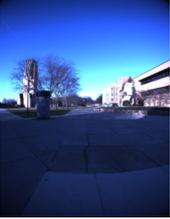 | 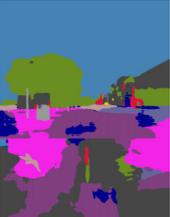 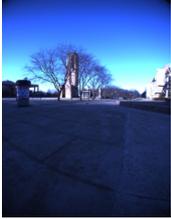 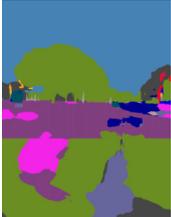 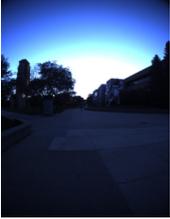 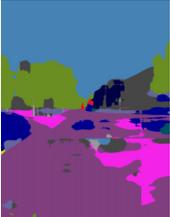 |
| 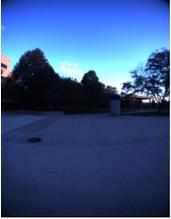 | 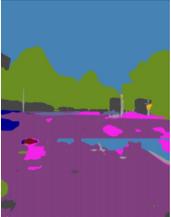 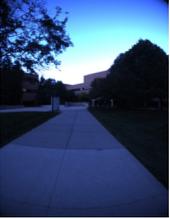 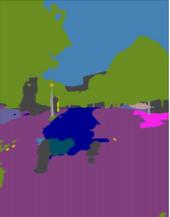 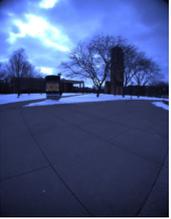 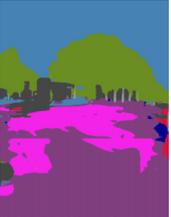 |
| 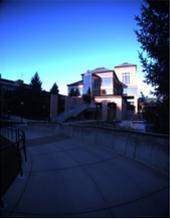 | 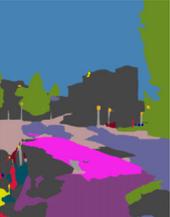 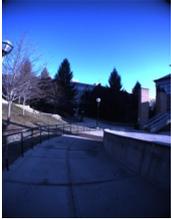 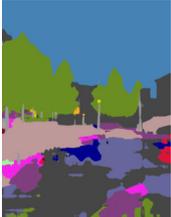 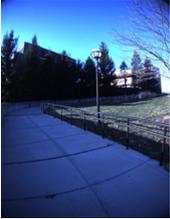 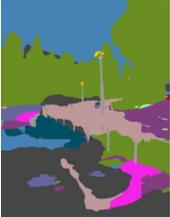 |
| 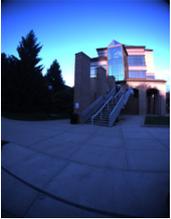 | 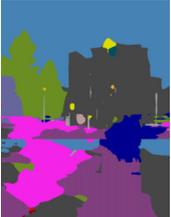 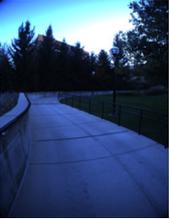 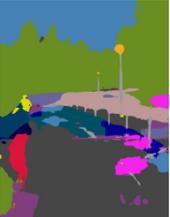 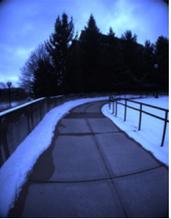 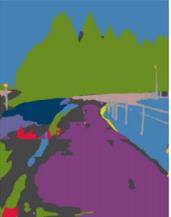 |
| 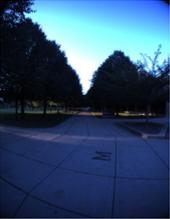 | 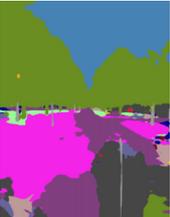 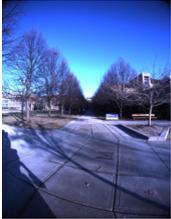 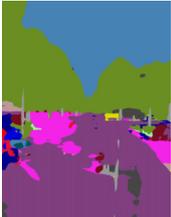 |
| 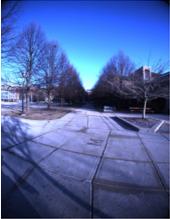 | 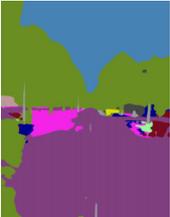 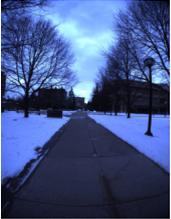 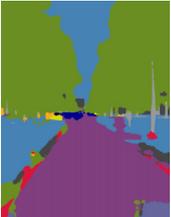 |

Figure 6: Localization results for NCLT dataset. Left-most image depicts database scene, while images to the right show successful localization results under different viewpoints and illumination/season. Top row shows RGB images while bottom row shows corresponding semantic segmentation. Results continued on next page in Figure 7.

| Database | Successful Queries |
|---|---|
| 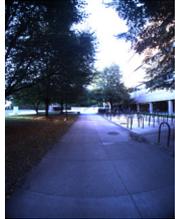 | 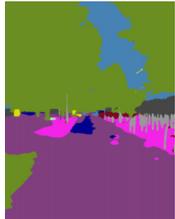 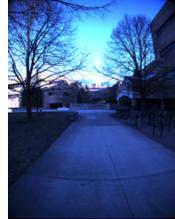 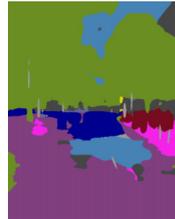 |
| 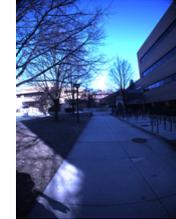 | 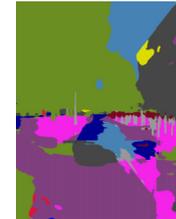 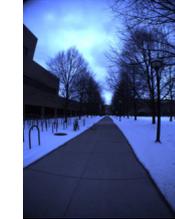 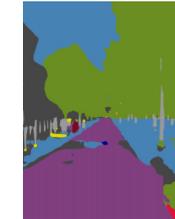 |
| 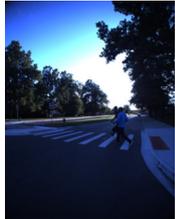 | 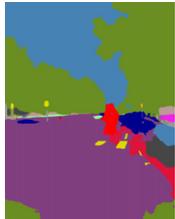 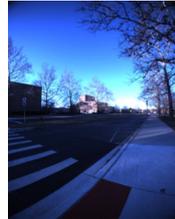 |
| 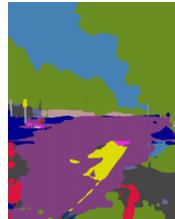 | 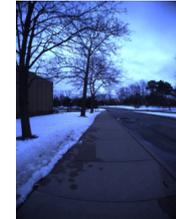 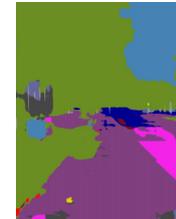 |
| 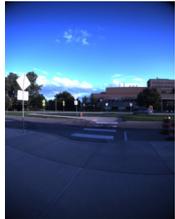 | 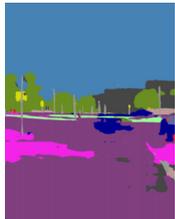 |
| 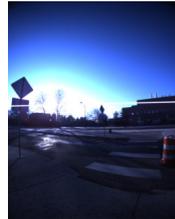 | 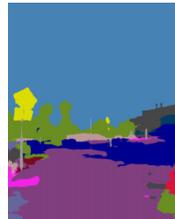 |

Figure 7: More localization results for NCLT dataset, continued from Figure 6.